# A Deep Learning Framework for Spatiotemporal Ultrasound Localization Microscopy

Léo Milecki, Jonathan Porée, Hatim Belgharbi, Chloé Bourquin, Rafat Damseh, Patrick Delafontaine-Martel, Frédéric Lesage, Maxime Gasse, Jean Provost

**Abstract**—Ultrasound Localization Microscopy can resolve the microvascular bed down to a few micrometers. To achieve such performance microbubble contrast agents must perfuse the entire microvascular network. Microbubbles are then located individually and tracked over time to sample individual vessels, typically over hundreds of thousands of images. To overcome the fundamental limit of diffraction and achieve a dense reconstruction of the network, low microbubble concentrations must be used, which lead to acquisitions lasting several minutes. Conventional processing pipelines are currently unable to deal with interference from multiple nearby microbubbles, further reducing achievable concentrations. This work overcomes this problem by proposing a Deep Learning approach to recover dense vascular networks from ultrasound acquisitions with high microbubble concentrations. A realistic mouse brain microvascular network, segmented from 2-photon microscopy, was used to train a three-dimensional convolutional neural network based on a V-net architecture. Ultrasound data sets from multiple microbubbles flowing through the microvascular network were simulated and used as ground truth to train the 3D CNN to track microbubbles. The 3D-CNN approach was validated *in silico* using a subset of the data and *in vivo* on a rat brain acquisition. *In silico*, the CNN reconstructed vascular networks with higher precision (81%) than a conventional ULM framework (70%). *In vivo*, the CNN could resolve micro vessels as small as 10 $\mu m$ with an increase in resolution when compared against a conventional approach.

*Index Terms*—Deep Learning, Ultrasound Localization Microscopy.

## I. Introduction

Ultrasound Localization Microscopy (ULM) bypasses the intrinsic spatial resolution of conventional contrast-enhanced ultrasound imaging via the localization of sparse microbubbles (MB) populations across ultrasound images [1], [2], [3]. As of today, ULM appears to be the only cost-effective, non-invasive, and non-ionizing method for the imaging of the microvasculature in large fields of view in vivo and in several organs such as the brain [4]. Adding tracking algorithms to the detection of MB enabled to map blood flow velocity maps [5]. The study of the microvascular angioarchitecture and its function at-depth and *in vivo* could become a powerful tool in the development of novel biomarkers for neurodegenerative diseases, cardiac diseases and cancer [6]. Nevertheless, the clinical application of ULM is limited essentially by its required long, motion-free acquisition time (a few minutes) to output a single highly resolved image.

This issue can be addressed in part by increasing MB density [7]. However, higher densities increase the difficulty of precisely localizing MB. Indeed, as they flow throughout the vascular network, MB that are close to one-another lead to US signal interference, preventing their accurate localization with a peak detection algorithm. Several processing techniques have been proposed to tackle this multi-object localization in ultrasound images. In [8] and [9], efficient filtering methods have been introduced based respectively on background removal, spatio-temporal-interframe-correlation based data acquisition, and separating spatially overlapping MB events into sub-populations. Some also use advance pairing techniques that discard unrealistic MB trajectories [10] or graph-based MB tracking on denoised images [11]. In [12], authors exposed the encouraging capacity of neural networks to spatio-temporally filter single MB in in vivo ULM images by training CNN to perform conventional signal processing methods. Others investigated the application of deep learning-based algorithms to enhance the localization of individual MB when higher concentrations are used. Those were either based on radiofrequency (RF) data [13] or envelope-detected images [14], [15], [16], [17] and all relied on a per-frame localization.

Manuscript sent July 01, 2020. This work was supported in part by the New Frontiers in Research Fund under Grant NFRFE-2018-01312, the Canadian foundation for innovation, John R. Evans Leaders Fund – Funding for research infrastructure under Grant 38095, Transmedtech, Ivado, and the Canada First Research Excellence Fund (Apogee/CFREF).

L. Milecki was with the Department of Engineering Physics, Polytechnique Montréal, Montréal, QC, Canada. (e-mail: leo.milecki@centralesupelec.fr).
J. Porée is with the Department of Engineering Physics, Polytechnique Montréal, Montréal, QC, Canada. (e-mail: jonathan.poree@polymtl.ca)
H. Belgharbi is with the Department of Engineering Physics, Polytechnique Montréal, Montréal, QC, Canada. (e-mail: hatim.belgharbi@polymtl.ca).
C. Bourquin is with the Department of Engineering Physics, Polytechnique Montréal, Montréal, QC, Canada. (e-mail: chloe.bourquin@polymtl.ca)
R. Damseh is with the Institute of Biomedical Engineering, Polytechnique Montréal, Montréal, QC, Canada. (e-mail: rafat.damseh@polymtl.ca)
P. Delafontaine-Martel is with the Department of Electrical Engineering, Polytechnique Montréal, Montréal, QC, Canada. (e-mail: patrick.delafontaine-martel@polymtl.ca)
F. Lesage is with the Department of Electrical Engineering, Polytechnique Montréal, Montréal, QC, Canada. (e-mail: Frederic.lesage@polymtl.ca)
M. Gasse is with the Mila, QC, Canada. (e-mail: maxime.gasse@gmail.com)
J. Provost is with the Department of Engineering Physics, Polytechnique Montréal, Montréal, QC, Canada and the Institute of Cardiology of Montréal, Montréal, QC, Canada. (e-mail: jean.provost@polymtl.ca)

More recently, authors proposed spatio-temporal filtering based on CNNs methods to localize multiple MB in in vitro data [18] or in small patches containing single MB to perform in vivo ULM [19]. Ideally, multi-object localization would include modelling complexities such as MB interferences, hemodynamic considerations, and blood vessels shape a priori.

In this study, rather than addressing the problem of sub-wavelength localization of individual MB in ultrasound images, we propose to resolve multiple MB trajectories from densely populated ultrasound cine loops. We hypothesized that the space-time ULM datasets carry richer information about the underlying microvascular network than individual ultrasound frames taken independently. Thus, we proposed a data-based approach as a substitute to the current localization task. Specifically, we demonstrated the capability of a 3D Convolutional Neural Network (CNN) to perform accurate MB tracking through ultrasound frames, which, when anchored in the ULM pipeline, enables efficient super-resolution microvascular imaging. To recover the microvascular network, we trained a 3D, V-net like CNN using ULM *in-silico* datasets based on a realistic microvasculature of a mouse brain extracted from *ex-vivo* 2-photon microscopy. The proposed framework was validated *in-silico* in independent data subsets and *in vivo* in a rat brain.

## II. METHODS

In this section, we first describe the conventional ULM pipeline used in this study and similar to other approaches described in the literature [7]. We then introduce the proposed novel Deep-stULM framework as a surrogate to the conventional localization and tracking step to recover dense tracks from ULM ultrasound image sequences.

### A. Ultrasound Localization Microscopy Pipeline

Following an injection of MB contrast agents, a programmable ultrasound scanner was used to acquire several hundred blocks of hundreds of compounded plane wave images acquired at thousand frames per second and repeated every second for minutes-long acquisitions. After in-phase-quadrature complex (IQ) image formation (i.e., beamforming) [20], tissue was removed using a singular value decomposition (SVD) clutter filter [21], [22], to recover signals from underlying flowing MB. The SVD threshold was heuristically set from what we use in our standard ULM process. Individual MB were then identified as bright local maxima, within correlation maps resulting from the correlation of the beamformed IQ images with the spatial impulse response (PSF) of the imaging system. MB were then precisely located through subpixel gaussian fitting. MB subpixel positions were accumulated over time on a finer grid corresponding to approximately one tenth of the wavelength, to recover the micro vessel density maps (see Fig. 1.A) & B)).

The precise identification and localization of MB may be altered by MB ultrasound signal interferences. Indeed, two or more MB interfering with one another may generate a MB-like signal, at a ghost location, that may be mistaken for a real MB. Such artefacts are usually addressed by lowering MB concentration, but at the expense of acquisition time, or using pairing algorithms that eliminate inconsistent MB trajectories [7].

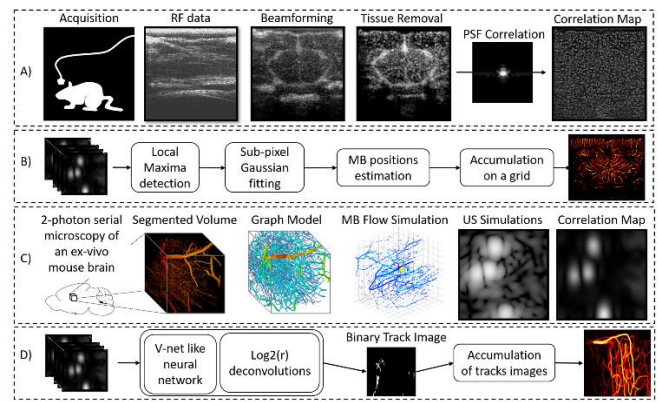

Figure 1 – Ultrasound Localization Microscopy (ULM) conventional and Deep-stULM framework: A) Ultrafast ultrasound in vivo acquisition and data normalization, B) Conventional sub-pixel localization and tracking procedure, C) ULM simulation framework and D) Proposed Deep Learning based MBs tracking procedure.

In this study, rather than locating individual MB in independent frames, and pairing them along the cine-loop to generate tracks, we designed a 3D-CNN that directly generates binary tracks out of the correlation maps cine-loops.

### B. Neural Network tracking model

The proposed tracking operation took $N_t$ successive frames of $N_z \times N_x$ correlation maps as input, i.e., $\chi \in \mathbb{C}^{N_t \times N_z \times N_x}$ with $|\chi_{ijk}| < 1 \; \forall (i,j,k) \in [\![1:N_t]\!] \times [\![1:N_z]\!] \times [\![1:N_x]\!]$ – depicting the flowing MB – , and produced one high-resolution binary tracking, i.e., $h(\chi; \theta) \in \{0,1\}^{r*N_z \times r*N_x}$ – depicting the 2D projection of the MB trajectories –, where $r$ is the resolution factor and $\theta \in \Theta$ is the set of parameters of the CNN architecture. We adopted a supervised learning setting and searched for the optimal tracking operation parameters $\hat{\theta}$ using a limited set of mappings $h(\cdot; \theta)$ and input/output pairs $(\chi, \psi) \in D$, the data set and by minimizing the empirical loss as follows:

$$\hat{\theta} = \underset{\theta \in \Theta}{\mathrm{argmin}} \sum_{(\chi,\psi) \in D} Loss(h(\chi; \theta), \psi) \qquad (1)$$

The fundamental idea behind our design was to attempt to detect microbubbles trajectories in spacetime rather than single points in space (e.g., the simplest case would be a 2D+t line representing a microbubble moving at a constant speed). The V-net part consisted of an encoder network which captured frames and temporal information into latent feature layers, and an expanding decoder network, which mapped this latent representation to detect MBs trajectories. A part of the model had to be destined to obtain high spatially sampled images. Thus, we designed a 3D CNN architecture as the concatenation of first, a fully-convolutional V-net-style architecture [23] followed by $log_2(r)$-deconvolution layers [24]. The original V-net architecture was modified. The number of down and up blocks was reduced from four to two in respect to our different input dimension and in order not to extremely reduce the spatial dimension in the feature space. Max pooling kernels value has been increased from two to four due to the high and unbalanced temporal dimension compared to spatial dimensions. Furthermore we did not reconstruct the temporal dimension in the decoder part (see Fig. 2) – as we aim to output a 2D image – but at the same time by not collapsing the temporal dimension to 1 too early to fully benefit from the 3D convolutions [25]. The second part of the network was aimed at increasing the spatial resolution. The $log_2(r)$ deconvolution layers were implemented through nearest neighbour resizing with a scale factor equal to 2 – so that output images size equal $r$ times input frames size – followed by a standard convolutional layer. The choice of using simple transposed convolutions instead has been evaluated and dissmissed, as it was producing checkerboards artifacts in generalization [26].

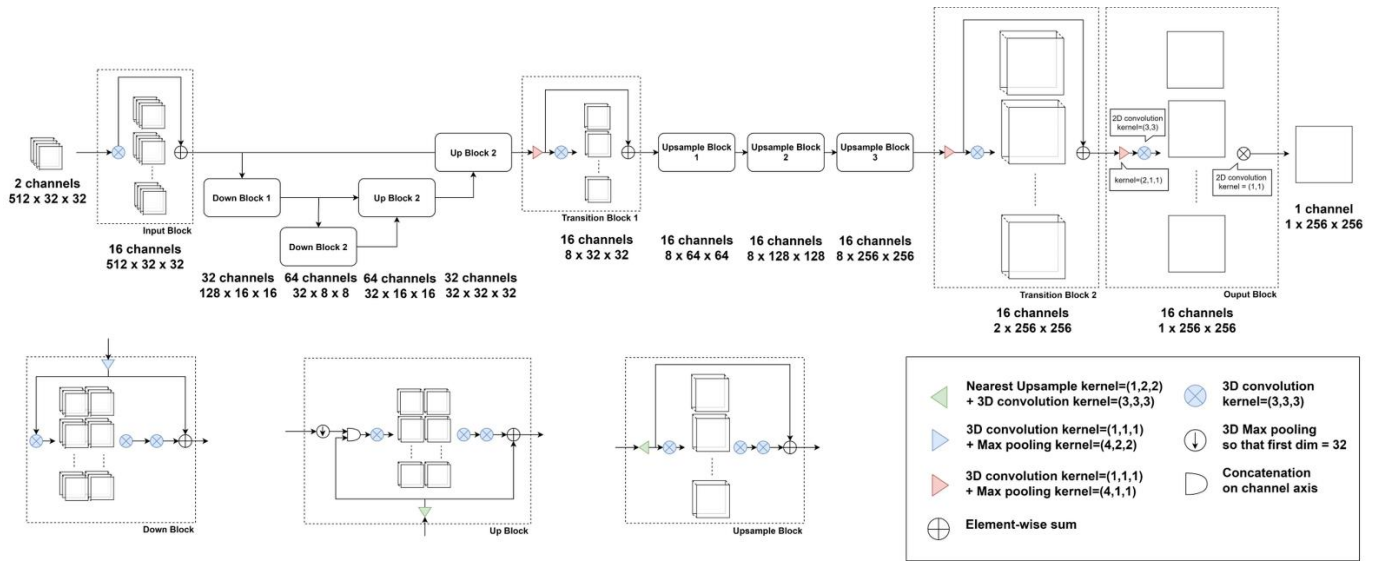

**Figure 2 - Neural network architecture of the 3D CNN.** In the first part, a V-net like autoencoder is used to detect MBs from a 2D+t ultrasound dataset and map tracks. In a second part, $log_2(r)$ deconvolution layers were used to increase the spatial sampling of those tracks.

Input correlation maps are complex images and since complex networks [27] are only at early stages of development, they were handled by feeding their real and imaginary parts as channels. The model was implemented using the PyTorch library and trained via the Adam optimizer [28] on a Nvidia GPU (GeForce RTX 2080 Ti). As the problem can be seen as an unbalanced pixel-wise classification task (tracking pixels represent a low percentage of the output image), we used the dice loss [29] after testing convergence speed over a few epochs against other binary classification oriented losses (negative log-likelihood, weighted negative log-likelihood and focal loss [30]).

To ensure and accelerate convergence, training was performed in two phases. The ground-truth binary tracks were dilated using a morphological dilation with a disk of radius 2 and were used as targets in the training of the 3D-CNN in the first phase. In the second phase, the primary ground-truth binary tracks were used as targets and the weights of the 3D-CNN were initialized as the weights obtained at the end of training in phase 1. The learning in two phases highly simplifies the optimization task and enabled a smaller final loss. In both phases, we used a multi-step learning rate scheduler which decays the learning rate of each parameter group by gamma once the number of epochs reaches one of the milestones. In phase 1, start learning rate was set to $10^{-1}$ with milestones=[15,45,75,100], gamma=0.1 and in phase 2, starting at $10^{-3}$ with milestones=[10,50,100], gamma=0.1. In both training, the total number of epochs was set to 150, which corresponds approximately to 48 hours per phase.

## III. TRAINING & EXPERIMENTS

To train the proposed CNN model, we designed an ULM simulation framework based on the microvascular network of a mouse brain [31] and ultrasound simulations [32].

### A. Microvascular network model

A graph-based model of a mouse brain microvasculature was first generated using two-photon microscopy as described previously [31]. Briefly, a fluorescent dye (i.e., dextran-FITC, 50 mg/ml in saline, Sigma) was injected through the tail vein of mice. 3D optical angiograms, with a $1.2 \times 1.2 \times 2 \ \mu m$ resolution, were then acquired using a two-photon microscope. Micro vessel segmentation was then achieved using a dedicated and retrained version of the FC-DenseNets semantic segmentation [33]. A surface mesh of the microvasculature was then computed using the marching cube algorithm. Finally, a reduced version of the generated surface model was contracted toward a 1D medial axis of the enclosed vasculature. The contracted mesh was then post-processed to generate a single-connected nodes graph with corresponding vessel radii [34]. To emulate realistic MB propagation throughout the microvascular network, Poiseuille flow conditions were simulated according to the vessel radii $R$ on the graph and average blood flow: $v_{mean}(R)$ as measured in rodent brain [35]:

$$v_{MB}(r,R) = v_{mean}(R) \cdot \left(1 - \frac{r^2}{R^2}\right) \qquad (2)$$

**Table I:** *In-silico & In-vivo acquisition parameters.*

| *Probe parameters* | | *Imaging parameters* | |
|---|---|---|---|
| **Center frequency** | 15.625 MHz | **Transmit frequency** | 15 MHz |
| **Number of elements** | 128 | **Waveform** | 3 cycles |
| **Linear Array pitch** | 0.1 mm | **PRF** | 3 kHz |
| **Element width** | 0.08 mm | **Compounding angles** | -1°,0°,1° |
| **Elevation Focus** | 8 mm | **Effective Frame rate** | 1 kHz |

Initial MB position were randomly distributed within the graph network, with a relative distance $r$ from the medial axis, and dynamically updated, with time, according to their relative velocity and position along the graph. Since vessels in our network ranged from 2 to 57.2 micrometers in diameter, maximum velocities ranged from 0.0093 mm.s$^{-1}$ to 5.4 mm.s$^{-1}$. MB concentration was varied, depending on the application, by randomly subsampling a single dense scatterer distribution.

### B. Ultrasound simulation model

To emulate the acoustic response of the MBs, we used an in-house GPU implementation of the SIMUS simulation software described in *[32]*. Specifically, SIMUS simulates the broadband acoustic pressure, recorded by every individual piezoelectric element, of an ultrasound linear array transducer, after plane wave excitation.

In this study, we simulated a 128-element, 15-MHz linear array transducer corresponding to the parameters of the probe (L22-14s Vermon, Tours, France) used in experiments. Simulation parameters are reported in Table I. Compounding angles were set to (-1°, 0°, 1°) according to our experimental setup and essentially limited by the data transfer rate of the acquisition system. The received pressure (corresponding to the radiofrequency (RF) data experimentally) was then subsampled into 100% bandwidth IQ channel data, in order to replicate the Verasonics Vantage system processing, and beamformed using an in-house GPU-implementation of the delay-&-sum beamformer [20]. For all experiments, IQ channel data were beamformed on a regular cartesian grid with $\lambda/4$ (*i.e.*, 25 $\mu m$) pixel size to prevent spatial aliasing.

### C. Training & evaluation dataset

We had access to six different anatomical sub volumes of $500 \times 500 \times 500\ \mu m$ in size: five were used to generate the training and validation set. To match the vessel size and density of a rat brain, as used for our *in vivo* experiments, anatomical sub-volumes were dilated by a factor of 2, resulting in 1-mm$^3$ sub-volumes [36],[37],[38]. IQ-data blocks of $32 \times 32$ pixels $\times$ 512 frames (corresponding to $800 \times 800\ \mu m \times 512\ ms$) were generated with a MB density at 5 MB.mm$^{-3}$. 500 such data blocks were generated for each volume, which lead to 2500 data blocks from which 90% were used for training and 10% for validation during the training process. The 6$^{th}$ sub volume, used neither for training nor validation, was used to generate the test sets composed of 800 data blocks (corresponding to a 14-min acquisition).

In the ULM method, the MB density parameter defines a trade-off between image quality (high densities hamper the precise localization of MB) and acquisition time (higher concentrations could lead to a reduced acquisition time, assuming all MB can be detected and are not obfuscated by interferences between MB). We wanted to evaluate the network capacity to perform tracking on different MB densities. We generated four test data sets simulating low MB density (1 MB.mm$^{-3}$), medium MB density (5 MB.mm$^{-3}$), high MB density (10 MB.mm$^{-3}$) and very high MB density (20 MB.mm$^{-3}$). In comparison, the recently proposed 2-D deep learning frameworks for ULM worked with a maximum of 3 MB.mm$^{-2}$ densities [13],[14].

According to [39], the theoretical resolution achievable with ULM with our current experimental setup is approximately 5 µm. Simulated IQ data-blocks with a spatial sampling $\frac{\lambda}{4} \times \frac{\lambda}{4}$ and the scaling factor $r = 8$ of our model led to output images at $\frac{\lambda}{4r} \times \frac{\lambda}{4r} \approx 3\mu m \times 3\mu m$ pixel size.

### D. Model validation

Validation of the CNN model on the tracking task was done by computing the average loss on the validation set, through training, and by calculating several classification scores on the validation set at the end of the training. Considering the tracking task as a pixel-wise binary classification problem, we based the validation on the confusion matrix: a table summarizing true positives, true negatives, false positives and false negatives (TP, TN, FP, and FN).

To evaluate the performance of the model we evaluated the precision (i.e., positive predictive value):

$$precision = \frac{TP}{TP + FP} \quad (7)$$

that quantify the probability of detecting an actual vessel among all detected vessels. We also evaluated the recall (*i.e.*, sensitivity):

$$recall = \frac{TP}{TP + FN} \quad (8)$$

that quantify the proportion of actual vessels that are correctly detected as vessels.

Finally, to provide a single measurement to discriminate our models' performance, we evaluated the Sørensen–Dice coefficient, that is the harmonic mean of the recall and precision.

$$Dice = \frac{2}{recall^{-1} + precision^{-1}} = \frac{2TP}{2TP + FP + FN} \quad (9)$$

The Dice coefficient quantify the similarity between two sets of data and allows, when the angiogram generated from all simulated MBs is taken as reference, to quantify the ability of a given ULM process to fill the entire network with MB. In this work, the Dice score is also referred as the network filling. In our preliminary studies, this first validation process enabled the selection of several model hyperparameters.

### E. From MB tracking to angioarchitecture images

The relevance of the proposed per data block approach appears through its insertion in the ULM pipeline (see Fig. 1). Indeed, the accumulation of MB tracks, similarly to the accumulation of estimated MB positions, can be used to compute a density mapping of the blood vessels. The inference was performed on these buffers and the binary tracking outputs were accumulated to recover the angiogram:

$$Angio = \sum_{k=0}^{N_{block}} h(\chi_k; \theta) \qquad (10)$$

In parallel, we applied the standard expert method previously described (see section II.A) on the same blocks for comparison. Because the proposed deep-stULM method generates binary tracks out of IQ-data-blocks while the standard method generates sub-wavelength positions, we generated pseudo tracks maps by accumulating these positions obtained through sub-pixel localization and without tracking on the predefined grid and binarized these density maps per data blocks. Such operation allows us to compare qualitatively the standard ULM process to the deep-stULM process, despite a partial loss of MB density information. Facing the difficulty to expose an accurate metric to compare angioarchitecture density maps, the quantitative results were measured through the evaluation of precision (5), recall (6) and dice (7) scores on the binarized density maps. Ground truth density maps were obtained from the accumulation of simulated MB positions on the latter predefined grid and then binarized to compute the proposed scores. The simulated vascular network was not directly used as ground truth as it could include vessels not perfused with MB and we do not intend our model to extrapolate the data and potentially generate spurious vessels.

### F. In-vivo acquisition

For *in vivo* experiments, we used MB concentrations reported in the literature [35]. While a direct conversion of these concentrations to number of microbubbles per mm$^3$ leads to very large values compared to our simulations (approximately 1000 MB.mm$^{-3}$), it has been shown that there is a two-order of magnitude reduction in MB concentration from *in vitro* to *in vivo* settings [40] which translates to ~10 MB.mm$^{-3}$ in our experiments. Furthermore, the tissue rejection process through Singular Value Decomposition and/or high-pass filtering is known to reject signals from slowly moving MBs [7]. It is thus difficult to determine the actual MB count flowing through the microvascular bed *in vivo*. To evaluate the impact of MB concentration *in vivo*, we designed a three-concentration experiments. Specifically, after craniotomy, increasingly concentrated MB solutions were injected in the tail vein (*i.e.*, 12.5 $\mu L$, 25 $\mu L$ and 50 $\mu L$, boluses of Perflutren Lipid Microsphere, Definity, Lantheus Medical Imaging, Billerica, MA, USA) prior to 5-minute-long ULM acquisitions. Each acquisition consisted of 384 RF data blocks of 400 compound plane wave images acquired at 1000 frames per second and repeated every second. Raw data were acquired with a Vantage system (Verasonics Inc., Redmond, WA) with the same parameters used for simulation (see Table II) and beamformed with a delay-&-sum beamformer. To calculate the correlation map from beamformed data, the PSF was simulated from the probe parameters and the acquisition sequence. The acquired Region of Interest (RoI) corresponded to a $512 \times 512$-pixel image beamformed at $\lambda/4$ resolution, which covers the entire brain (i.e., $\sim 13 \times 13$ mm RoI). To encompass the entire RoI with the 3D-CNN, we used a spatial sliding window. Specifically, the inference was done on spatiotemporal windows of $32 \times 32 \times 512$ pixels (i.e., the size of the 3D-CNN input) with a spatial overlap of 94 % (corresponding to a stride of 2 pixel in both directions). Overlapping regions were binary summed, which corresponds to a logical or between binary tracks regions. The $256 \times 256$ binary track outputs were cropped 32 pixels before and 32 pixels after in each spatial directions to avoid edge effects, coherently summed and binarized to form a single binary track image of the whole RoI per data block ($4096 \times 4096$ image). To consider organ displacement, we applied a rigid registration method between data blocks outputs, as described in [41].

## IV. RESULTS

### A. In silico performance of Deep-stULM

While varying the MB concentration, we can first study the effect of the number of data blocks (related to the acquisition duration), used in the ULM process, to recover the vessel network angiogram. Using ground truth simulated MBs, we analyzed the network filling ability of the ULM process (i.e., assuming that all MBs can be accurately located), as a function of the number of data blocks, for various MB concentrations (see Fig. 3) by taking all simulated MBs in the 6$^{th}$ vessel network used only for testing (i.e., 4000 data blocks @ 5 MB.mm$^{-3}$) to build a reference angiogram.

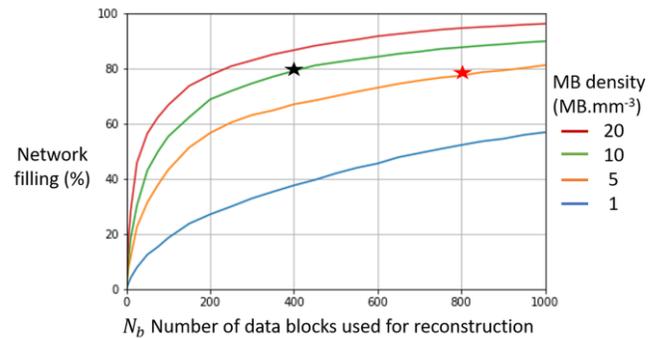

**Figure 3 - Vessel network filling as a function of the number of accumulated data blocks.** Dice score between the ground truth density map obtained with 4000 data blocks and the ground truth density map obtained with $N_b$ data blocks for different MB density.

The network filling capacity was quantified as the dice score between the ground truth angiogram (i.e., using all simulated MBs) and the intermediate angiograms computed with a fixed number $N_b$ of ground truth data blocks. This measure enables thereby to implicitly visualize the blood vessels' network filling against the acquisition time. We show in Fig. 3 that there is a saturation phenomenon as a function of the number of data blocks used for reconstruction. The higher the MB density, the earlier this saturation appears. In theory, we can achieve the same level of filling for half acquisition time when doubling the MB concentration. From these results, we can extrapolate the acquisition time necessary to achieve a complete reconstruction for a given concentration. Assuming all MB are accurately located, Fig. 3. gives the best-case scenario that can be obtained from our simulations. However, in vivo vessel network filling is subject to more variables than only time and more complex models should be considered in this case such as in [42], [43]. We performed ULM reconstructions with both standard ULM, here referred to as the expert method, and the proposed deep-learning framework, referred to as Deep-stULM (Fig. 4 and Table II) for various concentrations (i.e., 1 MB.mm$^{-3}$, 5 MB.mm$^{-3}$, 10 MB.mm$^{-3}$ and 20 MB.mm$^{-3}$) and for 800 data blocks (i.e., ~14 minutes acquisition).

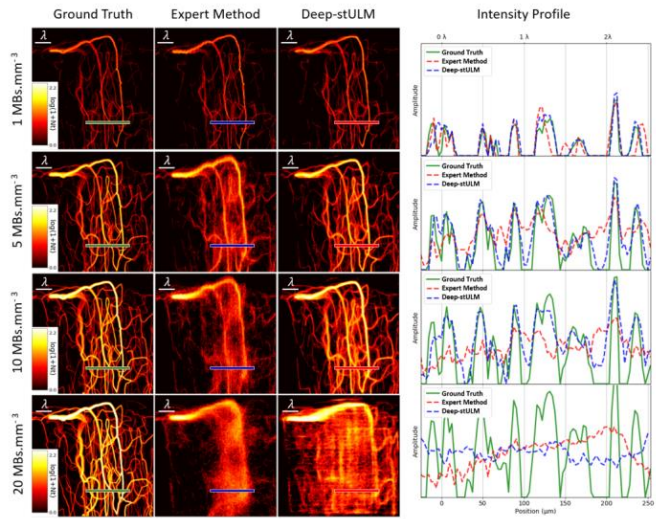

**Figure 4 – Effect of MB concentration on ULM angiogram reconstruction.** *In-silico* results comparing the ground truth (first column) to the prediction obtained by the proposed method (third column) and, for comparison, the considered expert model (second column) for the 4 data sets at different MB concentrations. The last column shows superposed intensity profiles along a given horizontal line. The dynamic of images is shown in $\log(1+N_{track})$, where $N_{track}$ is the number of tracks per pixel, to better discriminate low- and high-density regions.

**Table II: In silico performance of ULM angiogram reconstruction for an 800 data blocks acquisition @ different MB concentration compared to the angiogram generated from all simulated MBs (i.e., 4000 data blocks)**

| Concentration | Metrics (%) | Ground Truth | Expert Method | Deep-stULM |
|---|---|---|---|---|
| 1 MB.mm$^{-3}$ | Precision | 100 | 90 | **96** |
| | Recall | 35 | 33 | **40** |
| | Dice | 52 | 48 | **56** |
| 5 MB.mm$^{-3}$ | Precision | 100 | 74 | **87** |
| | Recall | 63 | 63 | **67** |
| | Dice | 77 | 70 | **78** |
| 10 MB.mm$^{-3}$ | Precision | 100 | 70 | **81** |
| | Recall | 78 | 64 | **82** |
| | Dice | 88 | 67 | **81** |
| 20 MB.mm$^{-3}$ | Precision | 100 | **67** | 58 |
| | Recall | 89 | 63 | **92** |
| | Dice | 94 | 65 | **71** |

For the lowest concentration (i.e., 1 MB.mm$^{-3}$), both methods showed comparable performance in terms of precision and recall. However, because of the low concentration, a network filling of only 48%, for the Expert method, and 56% for the Deep-stULM were obtained respectively. Larger concentrations enabled increased network filling, but at the cost of a loss in precision, which was more acute in the case of the expert method, especially for concentrations larger than 10 MB.mm$^{-3}$. Deep-stULM maintained high precision up to a 10 MB.mm$^{-3}$ concentration (81% vs 70% for the expert method) as showed in the intensity profiles (Fig. 4). At 20 MB.mm$^{-3}$, both methods failed at qualitatively resolving the angioarchitecture. At low MB concentrations (i.e., 1 & 5 MB.mm$^{-3}$) the proposed Deep-stULM framework outperformed the Ground truth in terms of recall and dice score (see Table II) which seems to indicate that the trained CNN can predict the presence of vessels from incomplete MB tracks.

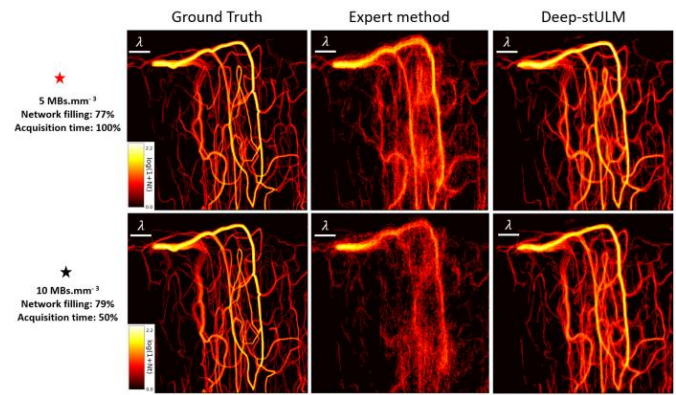

**Figure 5 – Effect of acquisition time on angiogram reconstruction.** *In-silico* results comparing the ground truth (first column) to the prediction obtained by deep-stULM (second column) for the 2 data sets at different concentration (i.e., 5 MBs.mm$^{-3}$ and 10 MBs.mm$^{-3}$) and equivalent total MB count.

We further evaluated whether it was possible to recover the vessel network in half the acquisition time when doubling the concentration from 5 to 10 MBs.mm$^{-3}$ (Fig. 5). The ground truth (left column) indicates that both scenarios can provide similar angiograms with a network filling of 77% and 79% respectively.

The expert model achieved network fillings of 70% and 60% respectively while Deep-stULM produced similar angiograms in both cases with a network filling close to the ground truth (i.e., 77% in both cases). Note, however, that the shorter acquisition time led to a reduction in precision: 87% down to 83% for the proposed Deep-stULM method and 74% down to 72% for the expert method. The reduction in precision and network filling is depicted in Fig. 5 by an increase of blurring artifacts and a loss in small vessel detection.

### B. In vivo performance of Deep-stULM

Fig. 6 shows an example of ULM reconstructions in a rat brain, after a 50-$\mu L$ bolus injection of microbubbles, using the expert method (bottom left panel) and the proposed Deep-stULM method (bottom right panel) compared to ultrasensitive Power Doppler with (top right) and without MB injection.

Both methods resolved the brain microvasculature with high contrast and sub-wavelength resolution. The Deep-stULM angiogram provided improved contrast and resolved smaller vessels that could not be detected with the expert method (see zoomed patches in Fig. 6).

To further analyze the impact of MB concentration in vivo on the proposed Deep-stULM capabilities, we recovered angiograms from the same animal, successively acquired for

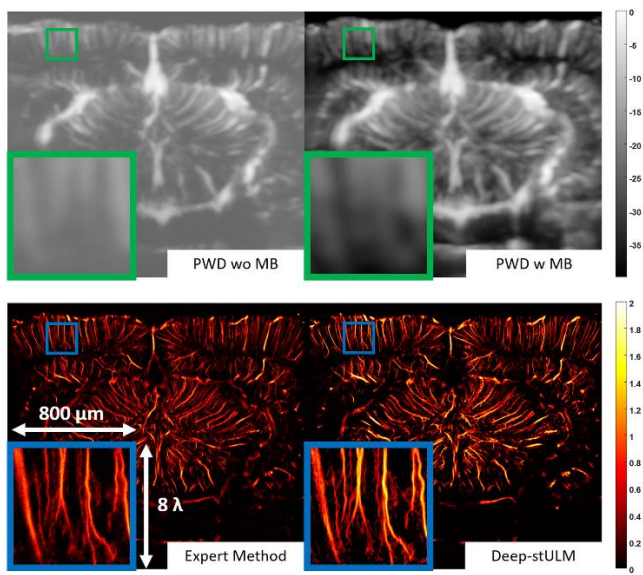

**Figure 6 - Ultrasound Localization Microscopy vs ultrasensitive Power Doppler in a rat brain:** Power Doppler without MB injection (top left panel), Power Doppler with MB injection (top right panel), Angiogram reconstruction using classical ULM processing (bottom left panel) and the Deep - stULM (bottom right panel) on a **50 μL** bolus ULM acquisition.

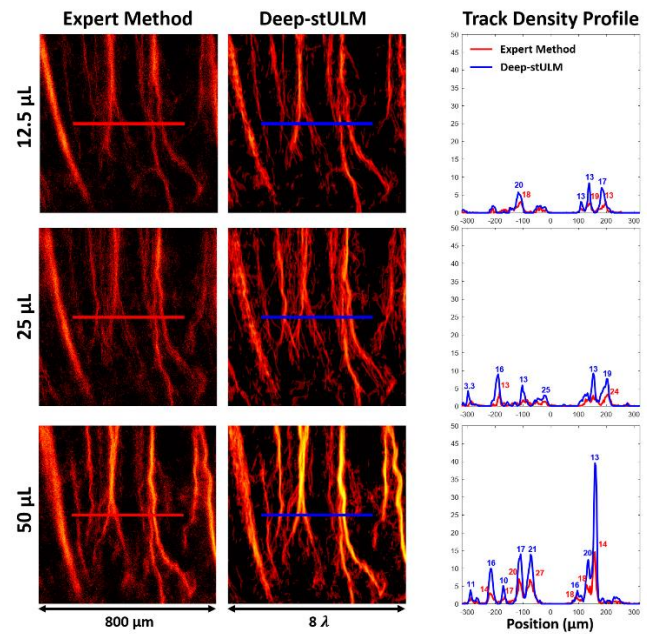

**Figure 7 – Effect of bolus concentration in vivo.** Zoom-in of ULM Angiograms using classical ULM processing (first column) and the Deep - stULM (second column) for three in-vivo dataset acquired successively after a 12.5-**μL** (first row), 25-**μL** (second row) and 50-**μL** bolus injections. The third column shows the track density profiles along a given horizontal line overlaid in red and blue with corresponding vessel diameters estimated as the Full Width at Half Maximum.

5 minutes after three bolus injections of 12.5 μL, 25 μL and 50 μL and waiting 5 minutes between acquisitions to make sure that the previous bolus was flushed out. Fig. 7 shows closeup patches and intensity profiles of the whole angiogram for the three concentrations. Overall, there was a good reproducibility between the acquisitions, even though they were taken five minutes apart. From the track density profiles showed in the zoom-in regions of Fig. 7, we can see that at low MB concentration (12.5 μL bolus), the number of well resolved vessels and their resolution are similar for both methods. While increasing the MB concentration, the number of well resolved vessels shows a difference as 6 vessels are detected by Deep-stULM whereas 2 are detected by the expert method within the 25 μL bolus experiment. At high MB concentrations (50 μL bolus), both methods demonstrate robustness to MB density as the numbers of peaks detected is similar. Nevertheless, the mean+/-std resolution over the 6 vessels detected by both methods is 16.1±3.5 μm for Deep-stULM and 18.3±4.1 μm for the expert method, with Deep-stULM resolving one more vessel at 11 μm. Overall, Deep-stULM enabled to resolve sufficiently more vessels than our expert method in comparison, especially small vessels that were less likely to be perfused. Looking at the track density profiles, the proposed Deep-stULM method appears to offer a better resolution than the expert model when increasing MB concentration and could resolve smaller micro vessels with diameters as small as 10 μm.

## V. DISCUSSION

In this study, we proposed a deep spatiotemporal convolutional neural network to recover subwavelength angiograms out of densely populated ULM acquisitions. Rather than locating/pairing multiple microbubbles in ultrafast ultrasound cineloops, the proposed Deep-stULM method was trained to resolve multiple MB tracks (i.e., trajectories) by considering both spatial and temporal information from the beamformed ultrasound datasets. The CNN was trained using an anatomically-realistic ULM simulation framework based on : highly-resolved (i.e., 2 $\mu m$) 3D angiograms acquired with two-photon microscopy in a mouse brain [31] and an in house GPU implementation of the SIMUS ultrasound simulation software described in [32] and validated both in silico, on a subset of the simulated data and in vivo on ULM acquisitions in a rat brain. In silico, we could achieve a two-fold reduction in the acquisition time, by doubling the microbubble concentration, while preserving the precision and specificity of super-resolved angiograms. In vivo, we obtained an increase in resolution as compared to standard ULM methods especially at high MB concentration.

MB separation and therefore precise localization in ULM mostly rely on the controlled MB concentration (e.g. using a continuous low-concentration MB perfusion instead of a bolus injection), to avoid MB signal interference, and minutes-long acquisitions, to fill the smallest vessels whose perfusion likelihood diminishes with size [35]. Microbubbles isolation in ultrasound images is a prerequisite to precise localization. If not achieved, the ULM process leads to a blurred and inconsistent reconstruction of the underlying network (see column 2 in Fig. 4). In this study, rather than isolating and locating individual microbubbles in independent frames, we trained a spatiotemporal CNN to isolate MB tracks in ultrasound cine-loops. Doing so, we hypothesized that overlapping/crossing MB tracks can be recovered even if they are interfering at some point in the sequence. To do so we adopted the well-known V - net architecture [23], originally designed for the segmentation of volumetric data in medical imaging, and trained it to recover highly resolved binary maps of multiple crossing MB tracks. With this simple framework, we were able to recover densely filled angiograms with very high precision where the standard ULM method collapsed (see Table II). This method paves the way to time-resolved ULM reconstruction through a significant reduction of the acquisition time.

The main objective of this study was to assess the capabilities of machine learning (ML) methods to recover vessel tracks from highly concentrated MB solutions circulating in a dense microvascular network. We demonstrated that the proposed CNN outperformed the now standard localization in terms of precision and network filling (see Table II).

The proposed Deep-stULM was extensively tested and validated both in silico and in vivo. Although in vivo results seemed promising and demonstrated the generalization capacity of the neural network, trained on in silico data, to achieve state-of-the-art results, several limitations deserve to be pointed out:

- Even though the correlation map sequence (i.e., resulting from the cross-correlation of the beamformed IQ data and the spatial impulse response of the imaging system) are complex data, they were treated as independent channels (i.e., a real channel and an imaginary channel) in the proposed CNN. Complex number CNN [27] would be better suited for this application. Indeed, from a signal processing point of view, using real and imaginary parts as separated channels through real number operations appears as a very indirect approach to solve this problem. The use of complex CNN would probably ease the optimization and convergence process while considering the instantaneous phase of the signal which is a well-known key feature in most ultrasound application (e.g. color Doppler).

- In the training process, we did not consider the underlying tissue signal which was assumed to be completely suppressed by the clutter filter process (i.e., Singular Value Decomposition). Even though the trained CNN performed well in vivo (see Fig. 6 & 7), it would highly benefit from more realistic training sets. One could, for example add incoherent white gaussian noise: to emulate acquisition noise, as well as coherent noise from real in vivo data to further train the CNN to distinguish MB signals from underlying tissue signals. Similarly, tissue motion could also be included in the simulation framework.

- The proposed method would probably perform better after data augmentation (i.e., translation, rotation, scaling etc…) to generate additional training datasets [44]. Coupled with the six vessel networks, the potential wide variability in datasets could enable us to better investigate reliability and robustness of our model through more complex training procedure such as tuning methods of hyperparameters or cross-validation between datasets. As an example, in this study, the CNN was trained with a fixed concentration of 5 MBs.mm$^{-3}$ which, from our experience seems to be close to the limit of the standard ULM method (see Fig. 4). Mixing different MB concentration in the training process would also benefit the proposed method without changing the model. One would have however to find the appropriate balance between concentrations. The richness of customization in this data simulation process paired with the proposed supervised learning method could be used as a powerful first step to transfer learning. Indeed, the future possibility to get access to annotated data in silico or even in vivo for the validation of ULM would be certainly limited in quantity, thus our model would be the starting point for the application of a supervised method on this gold standard data.

- Because our train sets only covers a 1 mm$^{-3}$ region of interest, we had to implement a sliding window algorithm to avoid Gibbs artifacts in the inference process in vivo. This drastically increased the inference time and might introduce unwanted artefacts. To bypass this problem, one could use data augmentation in the training process to cover wider RoIs. Data augmentation in this context must however be addressed with caution. Indeed, augmented data must remain consistent with a realistic microvascular network. Furthermore, in practice, the size of the input data block (i.e., 32 × 32 × 512 pixels) was originally chosen to fit on the physical memory of our graphic card (GeForce RTX 2080 Ti). Using larger RoIs is feasible in principle, but would require major changes in our approach and equipment.

- At low concentrations, the proposed CNN approach outperformed the ground truth in terms of network filling (i.e., Dice score) and recall. This suggests that the CNN interpolates blood vessels to locations where no or few MB have circulated. Note, however, that the training process was designed to prevent vessels extrapolation and the generation of undetected vessels, to ensure the validity of our vascular angioarchitecture maps, especially in presence of abnormalities. Exploring the capabilities of the proposed network to predict the existence of vessels in which no or few MB have circulated is the object of future work.

- In this study, the ULM process was formulated as a simple binary classification task (i.e., pixels inside or outside a vessel). Such classification was achieved through the minimization of a dice loss. However, beyond the main issue of MB separation, multiple close MB, which overlap or are crossing each other, would be considered as a single track in the target. Moreover, because of the complexity of the vascular network, one pixel may belong to more than one vessel (i.e., a bifurcation) or to a projection of several out-of-plane vessels. So, these sensitive regions could be subject to overfitting using the optimization of the dice loss function. To improve/generalize the classification process, one might consider using the generalized dice loss for multiclass segmentation as in [45]. Such a loss would most likely improve the specificity of the method in presence of a dense network.

- In ULM, detection of MB can enable, beyond vessels density information, to provide blood flow velocity information, for example by adding a pairing approach to the method [10]. But as in the proposed method, our model directly encodes spatio-temporal information to generate MB binary tracks, MB velocity information is lost. However, blood flow velocity is a highly relevant information toward several pathologies' indication and clinical purposes, as for example it is related to the efficient functioning of most organs. Thus, we would like to address this issue in future works. First, we would like to better investigate the impact temporal dimension size and sampling frame rate on the tracking performances. Then, as our model showed a good capacity to encode spatio-temporal information and as deep learning frameworks and more particularly losses are more suitable to predict probabilities information – labels, classes –, one could directly attempt to output clinically relevant classes which results from obtaining blood flow velocity information.

- In vivo results were obtained in the brain where a few minutes long motion-free acquisition is the most manageable. Our CNN model was trained on data without motion artefacts and in vivo inference per buffer enabled us to remove some blurry artefacts by rigid registration. However, chest organs imaging would require taking into account motion compensation in the model. Thus, the application of our proposed method to other organs would highly benefit from modifications consisting for example to insert motion compensation methods directly into the 3D-CNN model, as proposed in [46].

- The US data simulation and acquisition used in this framework have constrained us in the scope of this study to fix parameters such as the number of compounding angles or to heuristically set others such as the SVD threshold. Nevertheless, we intend to further investigate the impact of such parameters on the accurate tracking performance of the proposed method. Moreover, both the simulations and in vivo acquisitions were limited to the linear regime. Non-linearities associated to the microbubbles could potentially be leveraged within the proposed method to further enhance microbubble detection.

- In vivo results were analyzed through the comparison of Deep-stULM against our expert method. However, as to our knowledge there is no proper validation method for in vivo ULM, quality assessment of angioarchitecture images through criteria such as resolution or contrast highly suffers from not being in comparison to a ground truth image. Despite this constraint of not being disposed to introduce accurate metrics for whole images, we proposed a measure of resolution, which however introduced bias in manually selecting a region of interest and a few vessels in images containing hundreds.

## VI. CONCLUSION

We demonstrated the feasibility of recovering dense microvascular network angiograms, beyond the fundamental limit of diffraction, from highly concentrated ULM acquisitions using a Deep Convolutional Neural Network. The proposed CNN generalized well in vivo and offers a promising solution for time resolved ultrasound localization microscopy.